\documentclass[sigconf]{acmart}

\copyrightyear{2026}
\acmYear{2026}
\setcopyright{cc}
\setcctype{by}
\acmConference[MM '26] {Proceedings of the 34th ACM International Conference on Multimedia}{November 10--14, 2026}{Rio de Janeiro, Brazil.}
\acmBooktitle{Proceedings of the 34th ACM International Conference on Multimedia (MM '26), November 10--14, 2026, Rio de Janeiro, Brazil}
\acmISBN{979-8-4007-2213-4/2026/11}
\acmDOI{10.1145/3767308.3835259}

\begin{document}

\title{Geometry-Aware Camera Localization for Bronchoscopy}

\author{Lumin Chen}
\email{lumin.chen@cair-cas.org.hk}
\affiliation{%
  \institution{Centre for Artificial Intelligence and Robotics, Hong Kong Institute of Science \& Innovation, Chinese Academy of Sciences}
  \city{Hong Kong}
  \country{China}}

\author{Qingyao Tian}
\email{tianqingyao2021@ia.ac.cn}
\affiliation{%
  \institution{Institute of Automation, Chinese Academy of Sciences}
  \city{Beijing}
  \country{China}}

\author{Jinpeng Li}
\author{Haoyu Jiang}
\email{jinpeng.li@cair-cas.org.hk}
\email{haoyu.jiang@cair-cas.org.hk}
\affiliation{%
  \institution{Centre for Artificial Intelligence and Robotics, Hong Kong Institute of Science \& Innovation, Chinese Academy of Sciences}
  \city{Hong Kong}
  \country{China}}

\author{Huai Liao}
\author{Xinyan Huang}
\email{liaohuai@mail.sysu.edu.cn}
\email{hxinyan@mail.sysu.edu.cn}
\affiliation{%
  \institution{The First Affiliated Hospital, Sun Yat-sen University}
  \city{Guangzhou}
  \country{China}}

\author{Hongbin Liu}
\email{hongbin.liu@cair-cas.org.hk}
\affiliation{%
  \institution{Centre for Artificial Intelligence and Robotics, Hong Kong Institute of Science \& Innovation, Chinese Academy of Sciences}
  \city{Hong Kong}
  \country{China}}

\author{Dong Yi}
\email{dong.yi@cair-cas.org.hk}
\correspondingauthor
\affiliation{%
  \institution{Centre for Artificial Intelligence and Robotics, Hong Kong Institute of Science \& Innovation, Chinese Academy of Sciences}
  \city{Hong Kong}
  \country{China}}

\renewcommand{\shortauthors}{Lumin Chen et al.}

\begin{abstract}
Camera localization in bronchoscopy remains a challenging problem due to stringent accuracy requirements, real-time constraints, and limited training data. Compared to natural scenes, the confined anatomical structures demand millimeter-level precision, while intraoperative guidance necessitates low-latency inference. However, existing methods often fail to effectively exploit preoperative geometric priors, limiting their robustness and accuracy. To address these limitations, we propose a unified geometry-aware bronchoscope localization framework (GABL) that effectively fuses preoperative structural priors with paired intraoperative video to estimate 6-DoF camera poses. 
Specifically, to address visual ambiguity in complex airways, we propose a graph-guided coarse-to-fine localization scheme that effectively leverages structural priors for precise pose estimation. Furthermore, to mitigate pose jitter and bridge the visual-structural gap, we integrate a Transformer-based tracking model with a novel RGB-depth matching objective, jointly enforcing spatio-temporal and geometric consistency. 
Extensive experiments demonstrate that our method yields remarkable reductions of 8.37\% and 31.76\% in translation and rotation errors over the prior state-of-the-art, alongside 4 times inference speedup (33.6 FPS) for robust real-time bronchoscope localization. Project website:~\url{https://paulili08.github.io/GABL/}.
\end{abstract}




\begin{teaserfigure}
  \includegraphics[width=\textwidth]{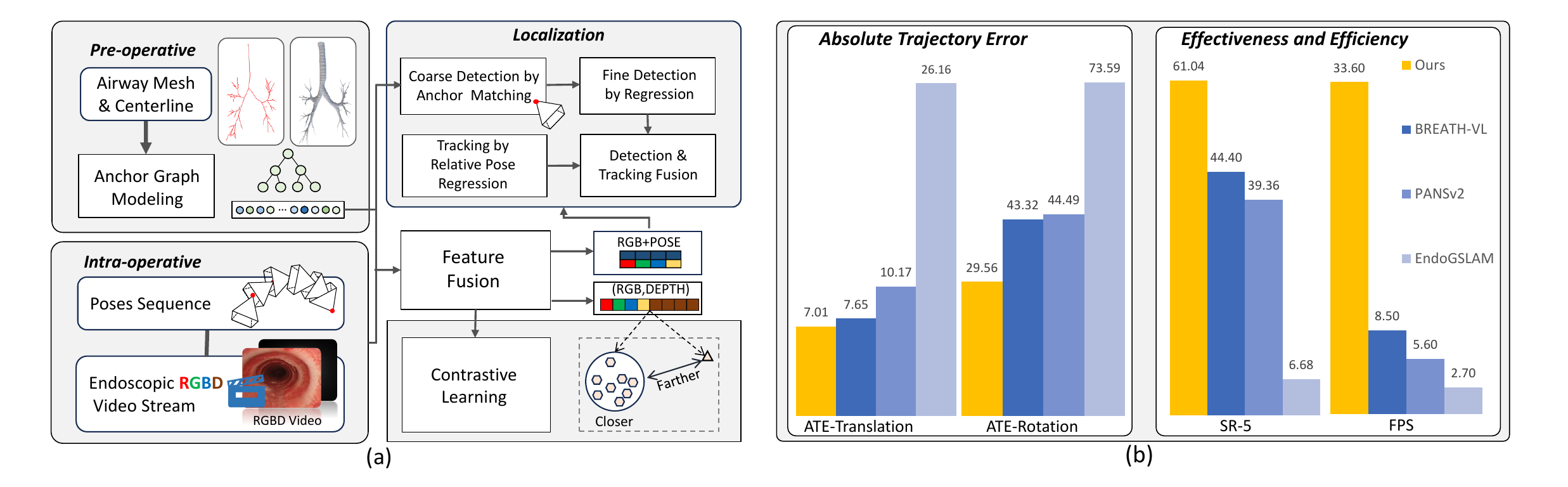}
  \caption{Overview and performance of Geometry-Aware Bronchoscopy Localization framework (GABL). (a) GABL combines preoperative airway geometry with intraoperative observations for anchor localization, temporal tracking, and RGB–depth matching. (b) Comparison with prior bronchoscopy localization methods in accuracy, success rate, and inference speed.}
  \label{fig:teaser}
\end{teaserfigure}


\maketitle

\section{Introduction}
Bronchoscopy localization aims to estimate the real-time 6-DoF pose of the endoscope within the airway, enabling accurate navigation and spatial awareness during minimally invasive procedures~\cite{ahn2020update}. This capability is critical for assisting physicians in reaching target regions, improving surgical safety, and enabling advanced applications such as image-guided intervention and augmented reality visualization~\cite{Review_endo}. However, achieving reliable localization in bronchoscopy is inherently challenging due to the complex and deformable anatomical structures, the prevalence of repetitive and low-texture surfaces, and the extremely limited field of view. These factors make it difficult to establish stable visual correspondences, while the small spatial scale of the airway further imposes stringent requirements on localization accuracy and real-time performance.

Despite significant progress in general camera localization, existing methods are not well suited for bronchoscopic scenarios. Most approaches rely on large-scale annotated datasets and visual features learned from natural images~\cite{lin2024exploring, wu2018image}, which suffer from severe domain gaps when applied to medical imagery. In addition, techniques designed for outdoor or indoor environments often assume rich textures, wide fields of view, and static scenes, conditions that do not hold in endoscopy. As a result, these methods tend to degrade significantly in the presence of ambiguous structures and dynamic tissue deformation. Consequently, directly applying general-purpose localization frameworks fails to meet the millimeter-level accuracy and robustness required in clinical bronchoscopy~\cite{cicenia2020navigational}.

To address these challenges, existing approaches primarily focus on exploiting geometric cues to compensate for the lack of reliable visual features in endoscopic imagery. On the one hand, one line of work reconstructs the anatomical structures using neural rendering techniques such as 3D Gaussian Splatting (3DGS)~\cite{wang2024endogslam}, aiming to build a consistent 3D representation of the scene from RGB observations. These methods often follow a Simultaneous Localization and Mapping (SLAM) paradigm, attempting to jointly reconstruct scene geometry and estimate camera trajectories, similar to pipelines developed for indoor environments. Another direction incorporates monocular depth estimation to provide additional geometric constraints for pose estimation, improving robustness in texture-sparse regions~\cite{sheikh2024dares}. On the other hand, several studies leverage preoperative imaging data, such as MRI or CT scans, as prior knowledge to guide intraoperative localization~\cite{tian2024pans, tian2025PANSv2, tian2026breathvl}. These priors can be represented in forms such as tree-structured surgical path graphs or deformable tissue meshes, enabling alignment between observed endoscopic views and patient-specific anatomical models. Overall, in the low-texture and low-light conditions characteristic of endoscopy, existing methods consistently seek to extract and utilize geometric information as the primary signal for camera localization, either from intraoperative RGB videos or preoperative priors. However, there are still issues with localization accuracy and inference speed that do not meet real-time requirements, as shown in Fig~\ref{fig:teaser}(b).

Based on the above observations, we aim to improve camera localization by systematically injecting geometric priors at three complementary scales: structure, motion, and appearance. Such multi-scale modeling has been shown to improve robustness and generalization in related tasks~\cite{wang2024mfrgn}. To this end, we propose a geometry-aware localization framework for bronchoscopic scenarios, which leverages both preoperative and intraoperative geometric information. \textbf{Anchor Localization}: We formulate camera localization as an anchor-based coarse-to-fine estimation problem. A graph representation of the airway is constructed from preoperative CT segmentation, where anatomical anchor points are encoded using a Graph Neural Network (GNN). The camera pose is progressively refined from anchor-level matching to precise alignment. \textbf{Temporal Motion Tracking:} To improve robustness, we model temporal dynamics using a causal Transformer, enforcing consistency across consecutive frames during training and inference. \textbf{Appearance-geometry Matching:} We introduce a cross-modal supervision between RGB observations and rendered depth images from the airway model, enabling geometry-consistent feature learning.

To instantiate this design, we present a Geometry-Aware Bronchoscopy Localization framework (GABL), which enforces explicit geometric supervision across all scales, as illustrated in Fig.~\ref{fig:teaser}. The main contributions are summarized as follows.

\begin{itemize}
    \item We propose a unified geometry-aware bronchoscopy localization framework (GABL) that systematically integrates geometry into localization at three scales, including anchor-based coarse-to-fine localization, temporal motion tracking, and appearance-geometry matching.
    
    \item We introduce an anchor-based graph representation derived from preoperative CT as structural geometric priors for bronchoscopy localization, and further incorporate depth-based representations from intraoperative RGB videos to provide complementary appearance-geometry supervision.

    \item We conduct extensive experiments on a clinically annotated bronchoscopy dataset, demonstrating that our method achieves best localization accuracy while maintaining real-time inference performance.
\end{itemize}

\section{Related Work}
\subsection{6-DoF Camera Localization}
Camera localization aims to estimate the full six-degree-of-freedom (6-DoF) pose of a camera, including its 3D position and orientation. Compared to object localization, this task requires jointly reasoning about translation and rotation, making it inherently more challenging. Existing approaches can be broadly categorized into feature-based methods, learning-based methods, and, more recently, neural rendering-based methods.

Feature-based methods formulate camera localization as a geometric matching problem between 2D image observations and a pre-built 3D scene representation~\cite{wu2018image}. These approaches typically rely on handcrafted local features to establish 2D–3D correspondences, followed by pose estimation using Perspective-n-Point (PnP)~\cite{Sarlin_2021_CVPR, 2022SCoRe, 2023COLMAP, moulon2016openmvg, Liu_2017_ICCV}. As a general-purpose solution, such methods achieve strong accuracy in well-textured and stable environments. However, they are sensitive to appearance variations such as illumination changes and motion blur, and often degrade in low-texture or repetitive scenes. In addition, the computational overhead of feature extraction and matching can limit efficiency in large-scale applications.

Learning-based methods estimate camera poses directly from images using deep neural networks, bypassing explicit feature matching and geometric solvers. These approaches can be broadly divided into two categories. Absolute pose regression methods feed images into convolutional or transformer-based networks~\cite{Kendall_2015_PoseNet, Kendall_2017_CVPR, Wang_2023_ICCV} to predict 6-DoF poses in an end-to-end manner. Scene coordinate regression methods instead predict dense 3D scene coordinates for each pixel~\cite{Brachmann_2017_DSAC, Guzman-Rivera_2014_CVPR, Giang_2024_CVPR, Yang_2019_ICCV}, which are then used in conjunction with RANSAC and PnP for pose estimation. By leveraging large-scale data, learning-based methods exhibit improved robustness to appearance variations compared to handcrafted features. However, they often suffer from limited generalization to unseen environments and may not achieve the same level of geometric accuracy as feature-based approaches. Recent efforts also explore incorporating additional geometric cues, such as depth or structure information, to further improve robustness.

Neural rendering-based methods have recently emerged alongside advances in 3D scene representation learning~\cite{wang2024nerfs, bao20253dgs}. These approaches model the scene as a continuous radiance field and perform camera localization via rendering-based optimization. Specifically, neural implicit representations encode spatial coordinates into color and density, enabling photorealistic novel view synthesis. Many methods follow a SLAM-like paradigm, where Neural Radiance Fields (NeRF) or 3D Gaussian Splatting (3DGS) are used for scene representation, while camera poses are iteratively optimized~\cite{rosinol2023nerf, matsuki2024MonoGS, ha2024rgbd, Murai_2025_CVPR, Yan_2024_CVPR}. Benefiting from strong scene modeling capacity, these methods demonstrate improved robustness in challenging conditions. However, they typically require substantial computational resources, rely on iterative optimization for pose estimation, and are often scene-specific, requiring significant time to adapt to new environments.

In addition, to address challenges in low-texture and low-light environments, graph-based representations have emerged as an effective form of geometric prior. By modeling a scene as a structured graph—where nodes represent keypoints, objects, or semantic regions, and edges encode spatial relationships—localization can be formulated as a graph matching problem, enabling geometric consistency to resolve ambiguities in repetitive structures~\cite{9665967, Sattler_2017_CVPR}. Furthermore, pose graph optimization refines camera trajectories by enforcing relative geometric constraints in SLAM systems~\cite{juric2021comparison}. Meanwhile, depth information provides direct geometric supervision. RGB-D SLAM methods leverage depth maps for scale-aware motion estimation~\cite{RGB_D_SLAM}, while recent approaches incorporate depth prediction networks and differentiable geometric losses to enforce both photometric and geometric consistency~\cite{zhan2021df_vo}. These developments highlight the importance of explicitly modeling geometric structure through either graph connectivity or depth cues for accurate and robust camera localization.

\subsection{Endoscope Localization}
With the rapid development of surgical robotics and endoscopic imaging systems, camera localization in minimally invasive surgical scenarios has attracted increasing attention~\cite{privitera2022image}. Compared to general-purpose environments, endoscopic localization presents several unique challenges. First, the surgical scene is dominated by deformable soft tissues that undergo continuous non-rigid motion during procedures, making it difficult to maintain consistent geometric correspondence over time. Second, the anatomical workspace is confined and safety-critical, requiring highly accurate and stable pose estimation~\cite{cold2024artificial}. Third, endoscopic images often suffer from challenging visual conditions, including limited illumination, specular highlights, and occlusions caused by fluids such as blood or mucus. Furthermore, the lack of distinctive texture in biological tissues significantly degrades the reliability of feature extraction and matching~\cite{chen2018slam}.

These factors collectively limit the effectiveness of conventional feature-based and learning-based localization methods, highlighting the need for approaches that better exploit both appearance and geometric information in dynamic and low-texture environments~\cite{ali2022we}.

Most existing datasets are derived from laparoscopic or colonoscopic procedures, such as SCARED~\cite{allan2021SCARED}, StereoMIS~\cite{hayoz2023StereoMIS}, EndoMapper~\cite{azagra2023endomapper}, EndoSLAM~\cite{ozyoruk2021endoslam}, and C3VD~\cite{bobrow2023C3VD}. Consequently, localization algorithms for these scenarios are relatively well developed. For example, EndoSLAM~\cite{ozyoruk2021endoslam} introduces Structure-from-Motion (SfM) into endoscopic localization. DARES~\cite{sheikh2024dares} leverages monocular depth estimation (e.g., Depth Anything v2~\cite{yang2024depth}) for self-supervised localization. EndoGSLAM~\cite{wang2024endogslam} incorporates 3D Gaussian Splatting (3DGS) with a streamlined representation to improve efficiency. Endo-FASt3r~\cite{sheikh2025endofast} adopts the foundation model Reloc3r~\cite{Dong_2025_Reloc3r} for pose estimation in endoscopic scenes.

In contrast, bronchoscopy localization remains relatively underexplored due to limited data availability. Deng \textit{et al.}~\cite{10342034Deng} construct a dataset from phantom lungs and ex-vivo human lungs, along with a feature-based visual odometry benchmark. PANS~\cite{tian2024pans} and PANSv2~\cite{tian2025PANSv2} introduce an in-vivo dataset collected from real bronchoscopic procedures and propose a landmark-based localization method. BREATH-VL~\cite{tian2026breathvl} incorporates visual-language models (VLMs) to enhance localization with semantic information and provides a dataset with preoperative CT and semantic annotations. Moreover, to address data scarcity and annotation challenges, ROOM~\cite{esposito2025room} proposes a framework for generating virtual bronchoscopy datasets from bronchial segmentation, offering a promising direction for data augmentation.

\section{Method}

\begin{figure*}[h]
  \centering
  \includegraphics[width=\textwidth]{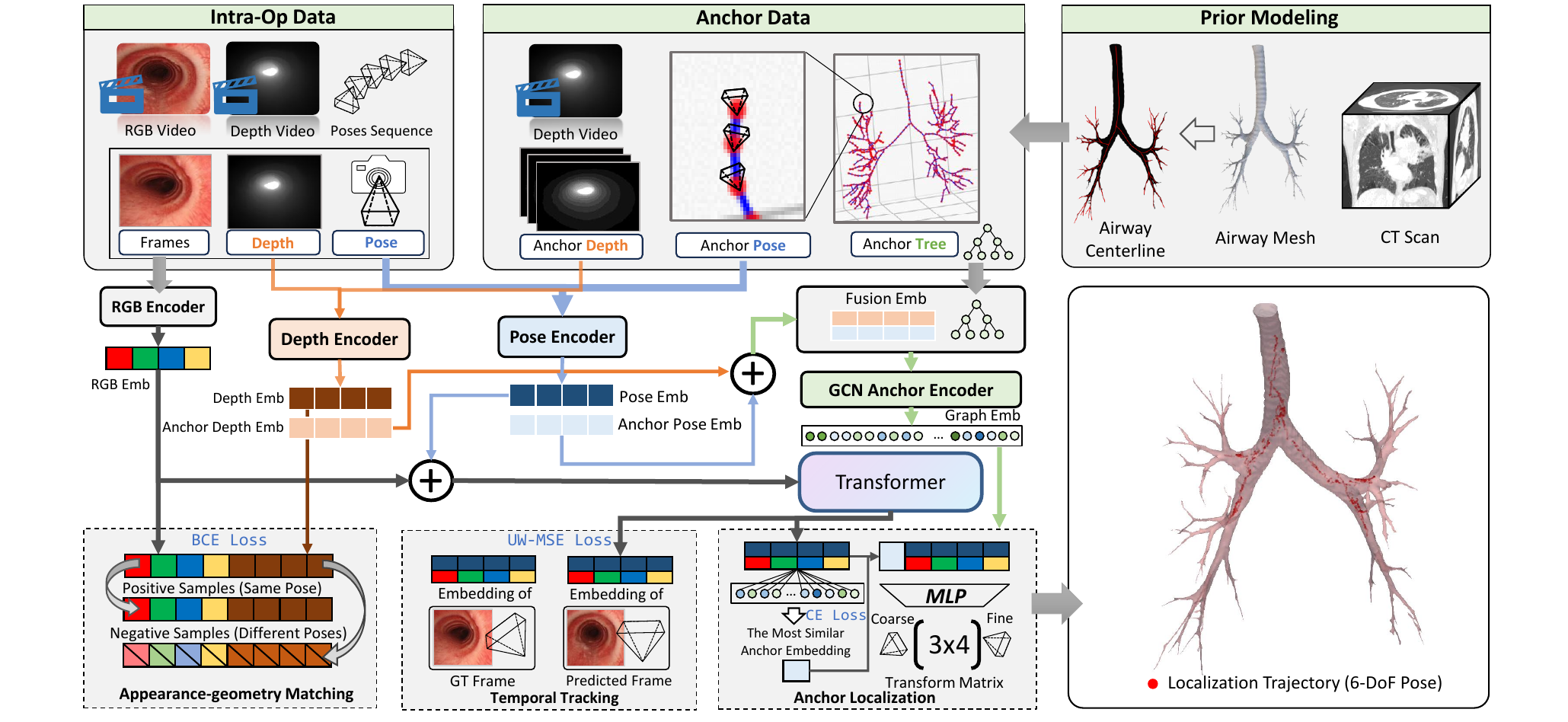}
  \captionsetup{width=\textwidth}
  \caption{Overview of GABL. The framework takes intraoperative data and anchor-based geometric priors derived from preoperative CT as inputs, and processes them with five representation encoders. It performs multi-stage localization via anchor-based coarse-to-fine estimation, temporal motion tracking, and appearance-geometry alignment, producing a 6-DoF camera trajectory.}
  \label{fig.2}
\end{figure*}

In this section, we present the proposed Geometry-Aware Bronchoscopy Localization framework (GABL). As illustrated in Fig.~\ref{fig.2}, the overall framework consists of three components: geometry prior construction from preoperative airway meshes, multi-modal representation learning of visual observations and geometric priors, and geometry-aware localization with unified supervision. The camera trajectory is obtained using the in-operative RGB video and the preoperative CT tracheal segmentation mesh.

We first construct a set of anatomical anchor-based geometry priors from the preoperative airway mesh, including a tree-structured representation, predefined 6-DoF poses, and pre-rendered depth maps. We then learn unified representations for both intraoperative observations and preoperative priors using modality-specific encoders, embedding RGB inputs and anchor-related information (e.g., graph structure, pose, and depth) into a shared feature space. Camera localization is performed within a unified framework via an anchor-based coarse-to-fine strategy, progressively refining pose estimates from anchor-level matching to precise alignment. To further improve robustness, we incorporate temporal motion modeling to enforce consistency across frames, and introduce an appearance-geometry alignment mechanism that aligns RGB features with rendered depth representations. All components are jointly optimized, enabling effective integration of structural, temporal, and cross-modal geometric supervision.

\subsection{3D Geometric Prior Modeling}
To improve training efficiency and reduce computational overhead, we first construct 3D geometric priors offline from preoperative CT images. We use the centerlines extracted from airway models in the dataset as the skeleton representation of the bronchi. Due to the narrow tubular structure of the airways, the centerline closely approximates the bronchoscope trajectory, providing an effective reference for subsequent localization. We uniformly sample a set of anchor points along the skeleton and construct an anchor graph based on the topological structure of the bronchial tree, ensuring coverage of key anatomical regions such as the carina and multi-level bifurcations.

After determining the anchor point coordinates, we generate corresponding camera poses to simulate the bronchoscope during navigation. Each camera pose is represented as a rigid transformation
\[
\mathbf{T} =
\begin{bmatrix}
\mathbf{R} & \mathbf{t} \\
0 & 1
\end{bmatrix}
\in SE(3),
\]
where $\mathbf{R} \in SO(3)$ denotes the rotation matrix and $\mathbf{t} \in \mathbb{R}^3$ represents the camera center. $SE(3)$ represents the space of rigid body transformations.

We fix the camera center $\mathbf{t}$ at each anchor point and define the target point as its adjacent node in the tree structure. Under this look-at constraint, the viewing direction is uniquely determined, which constrains two degrees of freedom of the rotation $\mathbf{R}$. As a result, the rotation space is reduced from three degrees of freedom to one. The remaining degree corresponds to a rotation around the viewing axis (roll), parameterized by an angle $\theta \in [0, 2\pi)$. We then render depth maps by projecting the 3D geometry under the given camera pose using a rasterization pipeline. Given the camera extrinsics and intrinsics, the depth at each pixel is defined as the distance along the viewing direction to the first visible surface, computed via z-buffering. Compared to Endo-FASt3r~\cite{sheikh2025endofast} and DARES~\cite{sheikh2024dares}, which rely on deep learning models for depth estimation, our approach eliminates the domain gap introduced by learned predictors while maintaining pixel-wise geometric consistency with the camera pose.

In the geometric prior modeling, we have constructed the anchor tree offline, generated camera poses for each anchor, and rendered the corresponding depth maps as geometric priors for supervision, thereby reducing computational cost during training.

\subsection{Representation Learning}
As illustrated in Fig.~\ref{fig.2}, we learn a unified embedding space for both intraoperative observations and preoperative geometric priors through a set of modality-specific encoders. The inputs consist of two groups: video observations, including RGB frames, corresponding camera poses, and depth maps; and anchor-related inputs, including the airway anchor graph, predefined anchor poses, and rendered anchor depth maps.

\textbf{Pose and Depth Encoding.}
We first encode geometric attributes shared by both inputs. Camera poses and anchor poses are encoded using a multi-layer perceptron (MLP) with three linear layers, producing pose embeddings and anchor pose embeddings, respectively. Similarly, depth maps from both domains are processed by a ResNetV2-18 backbone to obtain depth embeddings and anchor depth embeddings. This design enforces a consistent representation of geometry across intraoperative and preoperative domains.

\textbf{Visual Encoding.}
For intraoperative observations, RGB frames are encoded using a ResNetV2-50 backbone to extract visual features, resulting in RGB embeddings. These features capture appearance information complementary to geometric cues.

\textbf{Anchor Graph Encoding.}
To capture structural relationships among anchors, we represent the airway anchor tree as a graph $\mathcal{G} = (\mathcal{V}, \mathcal{E})$, where nodes correspond to anchor points and edges encode their topological connectivity. The graph is treated as undirected to enable bidirectional information propagation.

Each node is initialized with a feature vector by combining anchor pose and depth embeddings:
\begin{equation}
\mathbf{z}_i^{comb(0)} = \mathbf{z}_i^{\text{anchor pose}} + \mathbf{z}_i^{\text{anchor depth}},
\end{equation}

We then apply a three-layer graph convolutional network (GCN) to propagate information over the graph. At layer $l$, node features are updated as:
\begin{equation}
\mathbf{z}_i^{comb(l+1)} = \sum_{j \in \mathcal{N}(i)} \frac{1}{\sqrt{d_i d_j}} \mathbf{W}^{(l)} \mathbf{z}_j^{comb(l)},
\end{equation}
where $\mathcal{N}(i)$ denotes the neighborhood of node $i$, $d_i$ is the node degree, and $\mathbf{W}^{(l)}$ is a learnable transformation.

After three layers, we obtain the final anchor embeddings:
\begin{equation}
    \mathbf{z}_i^{\text{anchor}} = \mathbf{z}_i^{comb(3)},
\end{equation}
which encode both local geometric attributes and global structural context of the airway tree. This enables each anchor to aggregate information beyond its local neighborhood, capturing the hierarchical structure of the airway.

\textbf{Temporal Representation Learning.}
To incorporate motion dynamics, we employ a causal Transformer (pruned from Qwen3 backbone) to model temporal dependencies across video frames. The input to the Transformer is formed by combining RGB embeddings with pose embeddings. Let $\mathbf{z}^{\text{rgb}} \in \mathbb{R}^{n \times s \times d}$ denote the RGB embeddings for a video batch of size $n$ with sequence length $s$, and $\mathbf{z}^{\text{pose}} \in \mathbb{R}^{n \times s \times d}$ denote the corresponding pose embeddings.  

We introduce stochastic masking of pose embeddings as a regularization mechanism. First, a binary mask $\mathbf{m} \in \{0,1\}^{n \times s}$ is sampled from a Bernoulli distribution:
\begin{equation}
    \mathbf{m}_{i,t} \sim \text{Bernoulli}(p), \quad i=1,\dots,n, \ t=1,\dots,s,
\end{equation}

where $p=0.25$ is the probability of retaining the original pose embedding. The masked pose embeddings are then defined as:
\begin{equation}
    \tilde{\mathbf{z}}^{\text{pose}}_{i,t} = \mathbf{m}_{i,t} \cdot \mathbf{z}^{\text{pose}}_{i,t} + (1 - \mathbf{m}_{i,t}) \cdot \mathbf{z}^{\text{null}},
\end{equation}

where $\mathbf{z}^{\text{null}}$ is a learnable null vector representing a missing pose.

The masked pose embeddings are combined with RGB embeddings and fed into the causal Transformer:
\begin{equation}
\mathbf{z}^{\text{video}} = \text{Transformer}\big( \mathbf{z}^{\text{rgb}} + \tilde{\mathbf{z}}^{\text{pose}} \big),
\end{equation}

producing the final video embeddings $\mathbf{z}^{\text{video}}$ that encode both appearance and motion dynamics.

This stochastic pose dropout prevents the model from overfitting to precise pose cues and encourages robust temporal modeling based primarily on RGB observations, effectively regularizing motion dynamics learning.

Overall, this representation learning scheme bridges visual observations and geometric priors in a unified embedding space, providing the foundation for subsequent geometry-aware localization.

\subsection{Geometry-aware Localization with Unified Supervision}

\textbf{Anchor Localization.}
Our model adopts a two-stage anchor-based localization strategy. The coarse localization stage identifies the most relevant anchor as an initial pose estimate, followed by a fine localization stage for refinement.

Given an input frame, we extract its video embedding as  $\mathbf{z}^{\text{video}}_{i}$ from a context window. For a predefined set of anchor points, we denote their embeddings as $\{\mathbf{z}^{\text{anchor}}_{k}\}_{k=1}^{K}$.

We compute the similarity logits between the frame embedding and all anchor embeddings:
\begin{equation}
s^{\text{coarse}}_{i,k} = \left\langle \mathbf{z}^{\text{video}}_{i}, \mathbf{z}^{\text{anchor}}_{k} \right\rangle,
\end{equation}
where $\langle \cdot, \cdot \rangle$ denotes the inner product.

Let $y_i$ denote the ground-truth anchor label for frame $i$. The coarse localization is formulated as a classification problem using the cross-entropy loss:
\begin{equation}
\mathcal{L}_{\text{coarse}} = - \sum_{i} \log \frac{\exp(s^{\text{coarse}}_{i,y_i})}{\sum_{k=1}^{K} \exp(s^{\text{coarse}}_{i,k})}.
\end{equation}

The predicted anchor index is obtained as:
\begin{equation}
\hat{k}_i = \arg\max_{k} \ s^{\text{coarse}}_{i,k},
\end{equation}
and the corresponding anchor embedding is taken as the coarse pose representation:
\begin{equation}
\mathbf{z}^{\text{coarse}}_{i} = \mathbf{z}^{\text{anchor}}_{\hat{k}_i}.
\end{equation}

For fine-grained localization, the model predicts a refined pose $\hat{\mathbf{T}}_i$ by the Pose Regressor, along with a log-variance term $\log \sigma^2_{\text{fine}, i}$ that models the uncertainty of the prediction, following~\cite{Kendall_2018_uncertainty}.

We adopt an uncertainty-weighted mean squared error (MSE) as the training objective:
\begin{equation}
\mathcal{L}_{\text{fine}} = 
\frac{1}{N} \sum_{i}
\left(
\left\| \hat{\mathbf{T}}_i - \mathbf{T}_i \right\|_2^2 \cdot \exp(-\log \sigma^2_{\text{fine}, i})
+ \log \sigma^2_{\text{fine}, i}
\right),
\end{equation}
where $\mathbf{T}_i$ denotes the ground-truth pose.
This formulation allows the model to adaptively balance regression accuracy and prediction uncertainty.

The proposed two-stage anchor-based localization framework provides several key advantages by effectively leveraging geometric priors. First, the coarse localization stage transforms the continuous pose estimation problem into a discrete anchor classification task, which significantly reduces the search space and stabilizes training. By grounding predictions on predefined anchor points distributed along the airway skeleton, the model benefits from strong global geometric priors. Second, the fine localization stage refines the coarse estimate to achieve precise pose prediction. This coarse-to-fine strategy decouples global localization and local refinement, enabling the model to handle large spatial variations while maintaining high accuracy. Third, the use of structured anchors introduces dense and continuous geometric supervision across the entire trajectory, in contrast to sparse landmark-based methods. This design ensures robust performance even in regions lacking distinctive anatomical landmarks. Finally, the integration of geometric priors at multiple stages improves both translational and rotational accuracy, while effectively mitigating error accumulation over long sequences. Overall, the proposed framework provides a principled and efficient solution for geometry-aware visual localization. 

\textbf{Temporal Motion Tracking.}
Due to the high frame rate of the bronchoscopic video, the motion between adjacent frames is typically small. Based on this observation, we design a temporal motion tracking module to enforce consistency over time.

Given the video embedding of the current frame $\mathbf{z}^{\text{video}}_{i}$, the Pose Tracker predicts a relative pose offset $\Delta \mathbf{T}_i$ along with a log-variance term $\log \sigma^2_{\text{track}, i}$. The tracked pose is obtained by composing the predicted offset with the previous pose:
\begin{equation}
\hat{\mathbf{T}}^{\text{track}}_i = \mathbf{T}_{i-1} \oplus \Delta \mathbf{T}_i,
\end{equation}
where $\oplus$ denotes pose composition in $SE(3)$.

We supervise the tracked pose using the ground-truth pose $\mathbf{T}_i$ with an uncertainty-weighted mean squared error loss:
\begin{equation}
\mathcal{L}_{\text{track}} =
\frac{1}{N} \sum_{i}
\left(
\left\| \hat{\mathbf{T}}^{\text{track}}_i - \mathbf{T}_i \right\|_2^2 \cdot \exp(-\log \sigma^2_{\text{track}, i})
+ \log \sigma^2_{\text{track}, i}
\right).
\end{equation}

This formulation enforces temporal consistency while allowing the model to adaptively handle motion uncertainty.

\textbf{Appearance-Geometry Matching.}
The appearance-geometry matching module aligns RGB and depth features, introducing intraoperative geometric supervision into the localization framework.

Given the RGB embedding of the current frame $\mathbf{z}^{\text{rgb}}_{i}$ and a sampled depth embedding from intraoperative depth map $\mathbf{z}^{\text{depth}}_{j}$, we compute the matching logits as $s^{\text{match}}_{i,j}$.

To construct supervision, we adopt a stochastic sampling strategy. With a probability of $50\%$, we select the corresponding ground-truth depth map ($j=i$) as a positive sample; otherwise, we randomly sample a different frame ($j \neq i$) as a negative candidate.

Instead of using hard binary labels, we define soft labels based on pose similarity:
\begin{equation}
y^{\text{match}}_{i,j} = \exp\left( - \left\| \mathbf{T}_i - \mathbf{T}_j \right\|_2^2 \right),
\end{equation}
where $\mathbf{T}_i$ and $\mathbf{T}_j$ denote the poses of the current frame and the sampled frame, respectively. This formulation assigns higher similarity scores to geometrically close frames and lower scores to distant ones.

The appearance-geometry matching objective is defined using binary cross-entropy:
\begin{equation}
\mathcal{L}_{\text{match}} =
- \sum_{i}
\left[
y^{\text{match}}_{i,j} \log \sigma(s^{\text{match}}_{i,j})
+ (1 - y^{\text{match}}_{i,j}) \log \left( 1 - \sigma(s^{\text{match}}_{i,j}) \right)
\right],
\end{equation}
where $\sigma(\cdot)$ denotes the sigmoid function.

These losses jointly supervise the RGB encoder, pose encoder, depth encoder, anchor graph encoder, and temporal transformer, encouraging consistency between visual appearance and geometric structure.

\textbf{Inference Strategy.}
During inference, the model takes only RGB frames as input. The RGB encoder and causal Transformer extract video embeddings, from which the Pose Detector and Pose Tracker independently estimate the camera pose. The detector provides a geometry-aware pose estimate but may exhibit frame-to-frame jitter because of sparse anchors, whereas the tracker produces smoother, temporally consistent estimates but may accumulate errors over time. We therefore select between the two predictions according to their discrepancy. If the discrepancy exceeds an adaptive threshold, the tracked pose is used; otherwise, the detected pose is selected and the threshold is reset. The threshold increases linearly during consecutive tracking steps, encouraging a timely return to detection and preventing long-term drift. This strategy balances geometric accuracy and temporal consistency for robust pose estimation.

\section{Experiment}

\begin{table*}[h]
  \caption{Performance of 6-DoF localization on Bronchoscope dataset.}
  \label{tab:result}
  \begin{tabular}{lcccl}
    \toprule
    Methods & $\text{ATE}_{\text{trans}}(mm)$ $\downarrow$ & $\text{ATE}_{\text{rot}}(deg)$ $\downarrow$ & SR-5($\%$) $\uparrow$ & SR-10($\%$) $\uparrow$ \\
    \midrule
    \textbf{3D Gaussian Splatting} & & & & \\
    \text{MonoGS~\cite{matsuki2024MonoGS}} & 28.80 & 105.17 & 0.34 & 0.68\\
    \text{EndoGSLAM~\cite{wang2024endogslam}} & 26.16 & 73.59 & 6.68 & 19.19\\
    \midrule
    \textbf{Depth-based Representation} & & & & \\
    \text{Endo-FASt3r~\cite{sheikh2025endofast}} & 22.65 & 79.15 & 0.39 & 1.79\\
    \text{VNB~\cite{banach2021VNB}} & 43.52 & 109.22 & 7.06 & 15.64\\
    \text{DARES~\cite{sheikh2024dares}} & 44.58 & 109.46 & 2.02 & 4.32\\
    \midrule
    \textbf{Landmark Detection} & & & & \\
    \text{PANSv2~\cite{tian2025PANSv2}} & 10.17 & 44.49 & 39.36 & 67.37\\
    \text{BREATH-VL~\cite{tian2026breathvl}} & 7.65 & 43.32 & 44.40 & 75.83\\
    \midrule
    \text{Ours} & \textbf{7.01} & \textbf{29.56} & \textbf{61.04} & \textbf{83.66}\\ 
    \bottomrule
  \end{tabular}
\end{table*}

\subsection{Datasets and Evaluation Metrics}
To evaluate the performance of our model, we conduct experiments on the bronchoscopy localization benchmark BREATH~\cite{tian2026breathvl}. This dataset contains 66 bronchoscopy procedures, comprising 148,926 frames with 6-DoF pose annotations, along with airway meshes reconstructed from preoperative CT scans.

We adopt Absolute Trajectory Error (ATE) and Success Rate (SR) as evaluation metrics. ATE is decomposed into translational and rotational components: ATE$_{\mathrm{trans}}$ measures the mean Euclidean distance between predicted and ground-truth camera positions in millimeters, while ATE$_{\mathrm{rot}}$ measures the mean angular difference between predicted and ground-truth rotations in degrees. We further report SR-5 and SR-10, defined as the percentages of frames with translational errors below 5 mm and 10 mm, respectively.

\subsection{Implementation Details}
For 3D geometric prior modeling, we apply farthest point sampling~\cite{farthest_point_sampling} to select 512 anchor points along the airway skeleton, and construct a directed graph based on the topology of the bronchial tree. For each anchor point, two rotation angles in the fixed pose are determined according to the local tangent direction of the skeleton, while the remaining rotation angle is randomly sampled to complete the pose definition. Given the constructed poses, we utilize PyTorch3D to render corresponding depth maps from the airway mesh.

During training, our GABL model is trained on two NVIDIA A100 GPUs, with a per-GPU batch size of 6. Each input video clip contains 64 frames. All input images are resized to $256 \times 256$. The Pose Encoder is implemented as a 3-layer MLP, while the Anchor Encoder is a 3-layer GCN. The RGB Encoder and Depth Encoder adopt ResNetV2-50 and ResNetV2-18 backbones, respectively. The causal Transformer is based on the Qwen3-0.6B architecture and is trained from scratch without using any pretrained weights. To reduce computational cost, we halve its attention-head dimension, hidden size, and feed-forward intermediate size, and reduce the number of Transformer layers from 28 to 24. The resulting Transformer contains 81.82M parameters. All neural network modules in our model use Sigmoid Linear Unit (SiLU) activation functions.

We optimize the model using the AdamW optimizer with a learning rate of $1\times10^{-3}$, and momentum parameters $\beta_1=0.9$ and $\beta_2=0.99$. The training losses $\mathcal{L}_{\text{coarse}}$, $\mathcal{L}_{\text{fine}}$, $\mathcal{L}_{\text{track}}$ and $\mathcal{L}_{\text{match}}$ have a weighting of 1:1:1:1. The model is trained for 100 epochs.

Across the 66 procedure videos in the dataset, we use 56 cases for training and the remaining 10 cases for testing. The split is performed at the case level to avoid data leakage between training and test sets. To improve data diversity, we employ two data augmentation strategies. When segmenting each input video into clips, we randomly skip 1 to 5 frames to simulate temporal discontinuity. Additionally, each clip has a $50\%$ probability of being temporally reversed, which enhances robustness to motion direction. 

\begin{figure}[t]
  \centering
  \includegraphics[width=0.5\textwidth]{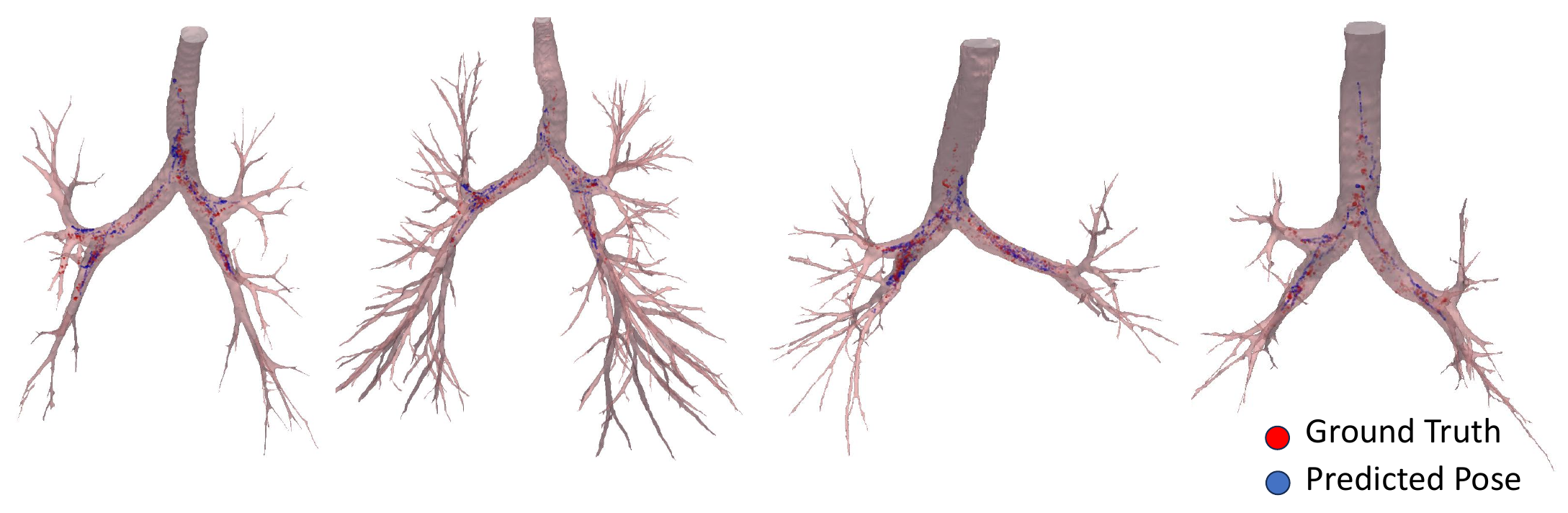}
  \captionsetup{width=0.5\textwidth}
  \caption{Visualization Results of Our Model}
  \label{fig.3}
\end{figure}

\subsection{Localization Results}
Table~\ref{tab:result} presents the localization performance on the BREATH dataset, comparing our method with existing state-of-the-art approaches. The compared methods can be broadly categorized into three groups based on their technical paradigms: 3D Gaussian Splatting (3DGS)-based methods, including MonoGS~\cite{matsuki2024MonoGS} and EndoGSLAM~\cite{wang2024endogslam}; depth-based representation methods, such as Endo-FASt3r~\cite{sheikh2025endofast}, VNB~\cite{banach2021VNB}, and DARES~\cite{sheikh2024dares}; and landmark-based methods, including PANSv2~\cite{tian2025PANSv2} and BREATH-VL~\cite{tian2026breathvl}.

As shown in Table~\ref{tab:result}, GABL achieves an $\text{ATE}_{\text{trans}}$ of 7.01 mm and an $\text{ATE}_{\text{rot}}$ of $29.56^\circ$, improving over the second-best method by $8.37\%$ and $31.76\%$, respectively. Its $SO(3)$ axis-wise rotation errors are $13.36^\circ$, $14.48^\circ$, and $15.39^\circ$. GABL also achieves SR-5 and SR-10 scores of $61.04\%$ and $83.66\%$. Qualitative results are shown in Fig.~\ref{fig.3}.

A closer examination of different method categories highlights the advantages of our geometric design. 3DGS-based methods tend to suffer from error accumulation due to the absence of explicit geometric constraints, leading to degraded localization accuracy over long trajectories. Depth-based methods introduce geometric cues via depth supervision; however, relying on depth alone provides limited structural guidance, which results in suboptimal performance in complex bronchial environments. Landmark-based methods improve localization by incorporating anatomical priors, such as airway carina or semantic landmarks. Nevertheless, these approaches rely on sparse or localized cues, which restrict their effectiveness in regions far from detected landmarks and limit their ability to provide consistent supervision along the full trajectory.

In contrast, our method leverages dense geometric priors by uniformly sampling anchor points along the entire airway skeleton. This design provides global structural coverage, continuous geometric supervision along the trajectory, and robustness in regions lacking distinctive landmarks. By integrating multi-scales of geometric constraints, our approach effectively reduces drift and improves both translational and rotational accuracy, leading to superior overall localization performance.

We evaluate the inference speed on a consumer-grade NVIDIA RTX 4090 GPU. Our model achieves 33.6 FPS, significantly outperforming PANSv2~\cite{tian2025PANSv2} and BREATH-VL~\cite{tian2026breathvl}, which run at 8.5 FPS and 5.6 FPS, respectively. In comparison, 3DGS-based methods such as EndoGSLAM~\cite{wang2024endogslam} exhibit lower efficiency in this scenario, achieving only 2.7 FPS in our experiments. This is likely due to the increased complexity of the airway lumen, which poses challenges for stable mapping and rendering. Overall, our method achieves a $4\times$ to $12\times$ speedup over prior approaches, while maintaining superior localization accuracy. This level of efficiency satisfies the real-time requirements of clinical endoscopy.

\begin{table}
  \caption{Ablation studies of localization components.}
  \label{tab:module}
    \begin{tabular}{l|cccc}
    \toprule
    \textbf{Settings} & $\text{ATE}_{\text{trans}}$ $\downarrow$ & $\text{ATE}_{\text{rot}}$ $\downarrow$ & SR-5 $\uparrow$ & SR-10 $\uparrow$ \\
    \midrule
    w/o GCN 
    & 10.67 & 31.48 & 57.59 & 78.55\\
    w/o Pose Regressor 
    & 9.31 & 107.54 & 41.09 & 77.09 \\
    w/o Pose Tracker 
    & 16.48 & 38.43 & 53.57 & 82.60 \\
    w/o Matcher 
    & 7.63 & 30.86 & 58.09 & 82.66 \\
    Full Model 
    & \textbf{7.01} & \textbf{29.56} & \textbf{61.04} & \textbf{83.66}\\
    \bottomrule
\end{tabular}
\end{table}

\begin{table}
\centering
\caption{Ablation study of data augmentation settings.}
\label{tab:data_ablation}
    \begin{tabular}{cc|cccc}
    \toprule
    \multicolumn{2}{c|}{\textbf{Settings}} & \multicolumn{4}{c}{\textbf{Metrics}} \\
    \midrule
    Skip & Reverse & $\text{ATE}_{\text{trans}}$ $\downarrow$ & $\text{ATE}_{\text{rot}}$ $\downarrow$ & SR-5 $\uparrow$ & SR-10 $\uparrow$ \\
    \midrule
    - & - & 7.97 & 31.71 & 54.75 & 80.72 \\
    \checkmark & - & 7.84 & 32.24 & 58.02 & 82.08 \\
    - & \checkmark & 7.71 & 33.56 & 59.65 & 81.52 \\
    \checkmark & \checkmark & \textbf{7.01} & \textbf{29.56} & \textbf{61.04} & \textbf{83.66}\\
    \bottomrule
\end{tabular}
\end{table}

\begin{table}
  \caption{Ablation study of dropout rate in causal Transformer.}
  \label{tab:dropout}
  \begin{tabular}{c|cccc}
    \toprule
    Dropout rate & $\text{ATE}_{\text{trans}}$ $\downarrow$ & $\text{ATE}_{\text{rot}}$ $\downarrow$ & SR-5 $\uparrow$ & SR-10 $\uparrow$\\
    \midrule
    0.25 & 8.04 & 31.51 & 57.94 & 83.12\\
    0.50 & 7.59 & 30.76 & 58.27 & 82.74\\
    0.75 & \textbf{7.01} & \textbf{29.56} & \textbf{61.04} & \textbf{83.66}\\
    1.0  & 7.94 & 31.25 & 58.57 & 82.48 \\
  \bottomrule
\end{tabular}
\end{table}

\subsection{Ablation Studies}
To validate the effectiveness of our geometry-aware design, we conduct ablation studies on four key components: the Anchor Encoder, Pose Regressor for fine-grained localization, Pose Tracker for temporal modeling, and Matcher for appearance-geometry alignment.

We first evaluate the contribution of each module by removing or modifying it individually. Specifically, we replace the GCN in the Anchor Encoder with a linear layer and remove the anchor tree input, thereby eliminating structured geometric priors from preoperative data. Removing the Pose Regressor means directly using coarse localization results as the final output. Disabling the Pose Tracker removes temporal motion supervision. Removing the Matcher prevents the model from learning the correspondence between RGB images and rendered depth maps, thus discarding geometry-aware representation learning.

As shown in Table~\ref{tab:module}, removing any component leads to performance degradation, demonstrating that each module contributes to the overall system. In particular, removing the Pose Tracker increases the mean $\text{ATE}_{\text{trans}}$ by 9.47 mm, indicating that temporal geometric constraints are essential to prevent error accumulation along the trajectory. 

Moreover, removing the Pose Regressor significantly degrades rotational accuracy, with $\text{ATE}_{\text{rot}}$ increasing to $107.54^\circ$. This suggests that coarse localization alone provides limited orientation information, and fine-grained geometric refinement is necessary for accurate pose estimation. 

When the anchor tree structure is removed and the GCN is replaced, the mean $\text{ATE}_{\text{trans}}$ increases to 10.67 mm. This result demonstrates that, beyond serving as coarse localization references, the tree-structured airway representation provides meaningful global geometric priors that improve localization accuracy.

As discussed previously, we employ two data augmentation strategies: random clip skipping and temporal reversal. We further analyze their impact in Table~\ref{tab:data_ablation}. These augmentations reduce the mean $\text{ATE}_{\text{trans}}$ by 0.70 to 0.96 mm. Random clip skipping exposes the model to temporally discontinuous inputs, improving robustness to irregular motion. Temporal reversal simulates backward navigation of the bronchoscope, enabling the model to better handle bidirectional motion patterns commonly observed in practice.

We further investigate the effect of pose dropout in the temporal Transformer. Specifically, we apply a Bernoulli distribution to randomly mask pose inputs for each frame, which prevents the model from over-relying on explicit pose signals. As shown in Table~\ref{tab:dropout}, a dropout rate of 0.75 achieves the best performance, indicating improved robustness to motion variations. In contrast, lower dropout rates result in a performance drop of 0.58 to 1.03 mm in $\text{ATE}_{\text{trans}}$, suggesting insufficient regularization. When the dropout rate is set to 1 (i.e., pose information is completely discarded), the model's performance declines, suggesting that combining pose information with RGB data is more effective for motion supervision. 

\section{Conclusion}
In this work, we present GABL, a geometry-aware localization framework for bronchoscopy that systematically integrates geometric priors at three complementary scales: structure, motion, and appearance. By combining anchor-based coarse-to-fine localization, temporal motion modeling, and cross-modal appearance-geometry supervision, our method unifies preoperative and intraoperative information within a single framework. Extensive experiments on the BREATH dataset demonstrate that GABL achieves a mean $\text{ATE}_{\text{trans}}$ of 7.01 mm and a mean $\text{ATE}_{\text{rot}}$ of $29.56^\circ$, with a success rate (SR-5) of $61.04\%$, outperforming existing methods while maintaining real-time inference performance. These results highlight the effectiveness of multi-scale geometric constraints for robust camera localization in challenging endoscopic environments and suggest promising potential for broader applications in medical and robotic navigation.

\begin{acks}
This work was supported by the InnoHK initiative of the Government of the Hong Kong Special Administrative Region.
\end{acks}

\bibliographystyle{ACM-Reference-Format}
\balance
\bibliography{sample-base}

\end{document}